\pdfoutput=1
\documentclass[11pt]{article}

\PassOptionsToPackage{table}{xcolor}

\usepackage[preprint]{acl}
\usepackage{times}
\usepackage{latexsym}
\usepackage[T1]{fontenc}
\usepackage[utf8]{inputenc}
\usepackage{microtype}
\usepackage{inconsolata}
\usepackage{booktabs}
\usepackage{graphicx}
\usepackage{adjustbox}
\usepackage{graphicx} 
\usepackage{multirow}
\usepackage{pifont}
\usepackage{caption}

\title{BiMediX: Bilingual Medical Mixture of Experts LLM}

\author{
Sara Pieri\thanks{*Equally contributing first authors.},
{\bf Sahal Shaji Mullappilly}\footnotemark[1],
{\bf Fahad Shahbaz Khan,} \\
{\bf Rao Muhammad Anwer,}
{\bf Salman Khan,}
{\bf Timothy Baldwin,}
{\bf Hisham Cholakkal} \\
\\
\textsuperscript{}Mohamed Bin Zayed University of Artificial Intelligence (MBZUAI) 
}

\begin{document}
\maketitle

\begin{abstract}
 In this paper, we introduce BiMediX, the first bilingual medical mixture of experts LLM designed for seamless interaction in both English and Arabic. Our model facilitates a wide range of medical interactions in English and Arabic, including multi-turn chats to inquire about additional details such as patient symptoms and medical history, multiple-choice question answering, and open-ended question answering. We propose a semi-automated English-to-Arabic translation pipeline with human refinement to ensure high-quality translations. We also introduce a comprehensive evaluation benchmark for Arabic medical LLMs. Furthermore, we introduce BiMed1.3M, an extensive Arabic-English bilingual instruction set covering 1.3 Million diverse medical interactions, resulting in over 632 million healthcare specialized tokens for instruction tuning. Our BiMed1.3M dataset includes 250k synthesized multi-turn doctor-patient chats and maintains a 1:2 Arabic-to-English ratio. Our model outperforms state-of-the-art Med42 and Meditron by average absolute gains of 2.5\% and 4.1\%, respectively, computed across multiple medical evaluation benchmarks in English, while operating at 8-times faster inference. Moreover, our BiMediX outperforms the generic Arabic-English bilingual LLM, Jais-30B, by average absolute gains of 10\% on our Arabic medical benchmark and 15\% on bilingual evaluations across multiple datasets. Our project page with source code and trained model is available at \url{https://github.com/mbzuai-oryx/BiMediX}  .
\end{abstract}

\begin{figure}[!t]
  \centering
    \includegraphics[width=\linewidth]{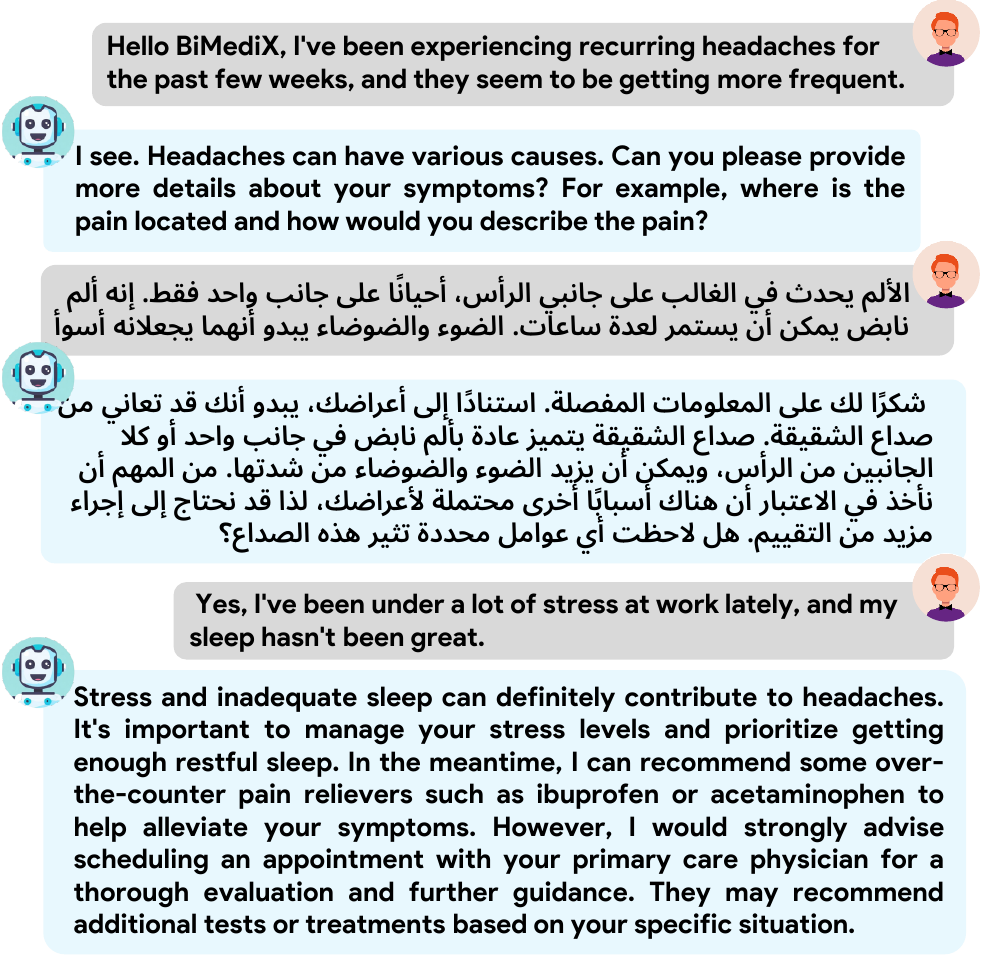}
    \caption{BiMediX is adept at conducting precise medical conversations seamlessly in both English and Arabic. It demonstrates advanced interactive capabilities, such as soliciting further details from the patient and seamlessly switching languages based on patient input while maintaining the continuity of the conversation.  
    The Arabic text translates as follows: \textbf{User:} \textit{The pain occurs mostly on both sides of the head, sometimes on only one side. It is a throbbing pain that can last for several hours. Light and noise seem to make it worse.} \textbf{BiMediX:} \textit{Thank you for the detailed information. Based on your symptoms, it sounds like you may have a migraine. Migraine headaches are usually characterized by throbbing pain on one or both sides of the head and can be triggered or increased in intensity by light and noise. It is important to consider that there are other possible causes for your symptoms, so we may need to do further evaluation. Have you noticed any specific factors that trigger these headaches?}}
    \label{fig:blingual_conv}
\end{figure}

\section{Introduction}
\begin{figure*}[!t]
  \centering
    \includegraphics[width=\linewidth]{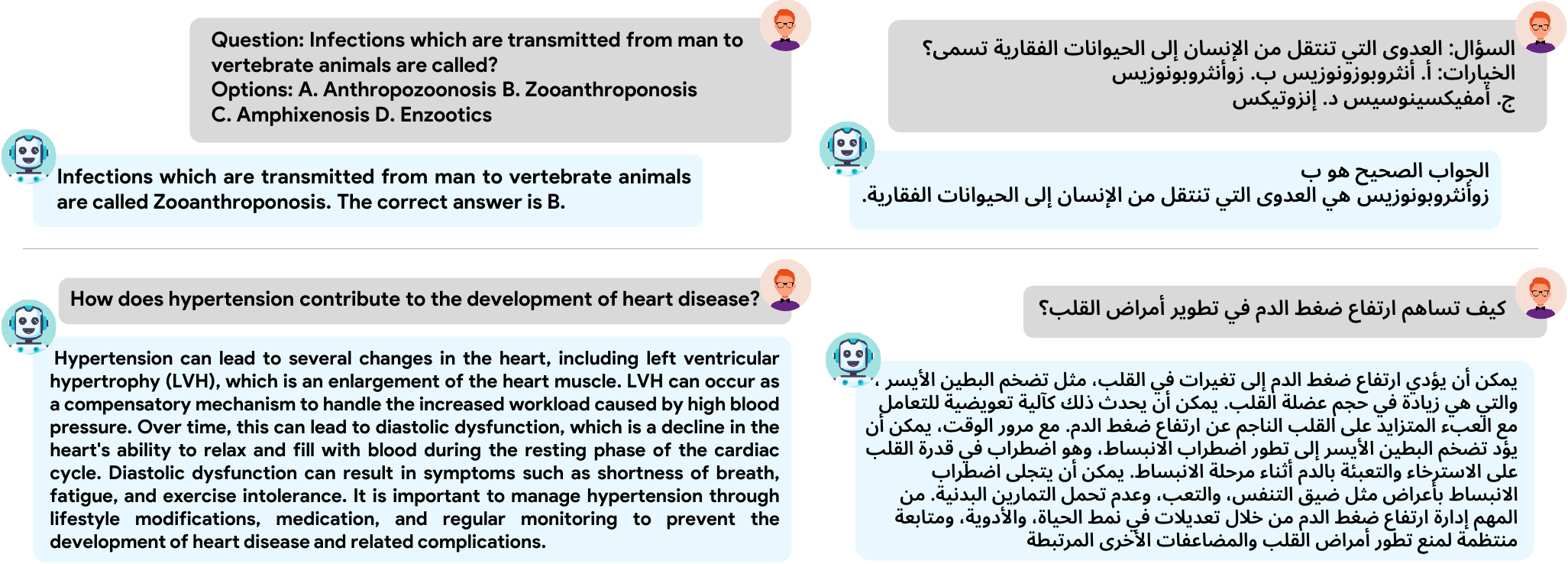}
    \caption{English and Arabic language capabilities of our BiMediX with multiple-choice questions and answers (MCQA) on the top row,  and open-ended questions and responses (QA) on the bottom row.}
    \label{fig:mcqa}
\end{figure*}

Large Language Models (LLMs) demonstrate excellent ability to understand instructions, generate human-like text, and tackle unseen or complex tasks, rendering them valuable for a broad range of applications. Specifically, LLMs have the potential to impact the medical field by offering virtual chat support across diverse medical departments to enhance diagnostic accuracy. However, despite the promise of LLMs in healthcare, the medical domain poses distinct challenges, including the necessity for domain-specific data and expertise to ensure accurate interpretations, as well as concerns regarding transparency and bias.

Recent methods have emerged to bridge the gap in medical domain knowledge by developing open-source medical LLM alternatives \cite{chen2023meditron, toma2023clinical, wu2023pmc} to proprietary models like ChatGPT \cite{achiam2023gpt}.  However, among the leading medical LLMs in English, Med42-70B \cite{med42} has not made its training data and resources available. 
Meditron-70B \cite{chen2023meditron}, relies on separate fine-tuning for each evaluation dataset. This approach is likely to compromise the model's ability to serve as a unified solution with diverse interaction capabilities.
Moreover, many of the leading open-source medical LLMs \cite{med42,chen2023meditron} are limited to the English language, posing limitations in their ability to understand and interact in resource-constrained languages such as Arabic. Despite Arabic's potential to cater to a population of more than 400 million people, it remains underrepresented in the medical LLM literature.   The unique features of Arabic, such as its distinctive script and less conventional right-to-left writing format, the unavailability of large-scale medical training data in Arabic, and the lack of a comprehensive benchmark to evaluate Arabic medical LLMs, present challenges in the development of Arabic or English-Arabic bilingual medical LLMs.
In this work, we introduce BiMediX, a bilingual medical mixture of experts LLM with seamless conversational capabilities in both English and Arabic. Our BiMediX is a chat agent with state-of-the-art performances on non-chat benchmarks in both English and Arabic.

\subsection{Contributions}
Our contributions can be summarized as follows: 

\noindent\textbf{(i)} We introduce the \textbf{\textit{first  bilingual medical mixture of experts LLM}}, named \textbf{\textit{BiMediX}},   having seamless interaction capabilities in both English and Arabic languages (see Fig.\ref{fig:blingual_conv}).   Our model facilitates various medical interactions, including multi-turn chats essential for follow-up inquiries with human patients (Fig.\ref{fig:blingual_conv}),  multiple-choice question answering (Fig.\ref{fig:mcqa} top row) and  open-ended question answering (Fig.\ref{fig:mcqa} bottom row).  

\noindent\textbf{(ii)} We develop a \textbf{\textit{semi-automated iterative translation pipeline}}, incorporating a human verification step to ensure high-quality translation of English medical text into resource-constrained Arabic. This pipeline facilitates the compilation of  instruction-tuning dataset and a comprehensive benchmark for evaluating Arabic healthcare LLMs and Arabic-English Bilingual LLMs.  The proposed Arabic evaluation benchmark can play a pivotal role in future research endeavors.

\noindent\textbf{(iii)} We curate a comprehensive \textbf{\textit{Arabic-English bilingual instruction set }} named \textbf{\textit{BiMed1.3M}} comprising over 1.3 million instructions, resulting in a total of over 632 million healthcare specialized tokens.  Our dataset  comprises open-ended question-and-answer, multiple-choice question answering, and over 200k synthesized multi-turn chats rooted in authentic medical content. The integration of multi-turn conversations allows for the creation of a chatbot capable of conducting follow-up inquiries with human patients. 
Our BiMed1.3M dataset maintains a 1:2 Arabic-to-English ratio across diverse medical interactions. 

\noindent\textbf{(iv)}   
We introduce a \textbf{\textit{parameter-efficient fine-tuning of routing and expert layers}}  in Mixtral \cite{jiang2024mixtral} using BiMed1.3M, requiring fewer training resources compared to Med42 and Chen et al. \cite{med42, chen2023meditron}.  Our BiMediX achieves \textbf{\textit{state-of-the-art performance  }} on multiple medical exam question datasets \textbf{\textit{in both English and Arabic}}. Our model outperforms Med42 and Meditron by average absolute gains of 2.5\% and 4.1\%, respectively, across multiple medical benchmarks in English, while operating 8 times faster. Additionally, BiMediX surpasses the generic Arabic-English bilingual  LLM, Jais-30B, by average absolute gains of 10\% on Arabic medical benchmarks and 15\% on bilingual evaluations across multiple medical datasets.  Moreover, bilingual instruction tuning with our BiMed1.3M  leads to an average absolute 10\% gain over the base mixtral \cite{jiang2024mixtral}  on bilingual evaluations.

\section{Related Works}

Proprietary models, including OpenAI's GPT-3.5 and GPT-4 \cite{achiam2023gpt}, as well as Google's Flan-PaLM \cite{chung2022scaling}, Med-PaLM \cite{singhal2022large}, and Med-PaLM 2 \cite{singhal2023towards}, have showcased remarkable achievements on English medical benchmarks.
Nonetheless, the scarcity of public information, alongside concerns regarding access and data privacy, motivate the development of open-source alternatives to proprietary models. These alternatives aim for customization, transparent evaluation, and safe clinical integration.
Early open medical language models, enhanced BERT-style models for medical purposes, leveraging the PubMed corpus for either continuous pre-training \cite{lee2020biobert, huang2019clinicalbert} or specific domain training from scratch \cite{beltagy2019scibert, gu2021domain, shin2020biomegatron}. Subsequent research enhanced bidirectional systems to incorporate link structures \cite{yasunaga2022linkbert} and knowledge graphs \cite{yasunaga2022deep}. Building on the GPT model series, later studies have tailored GPT-2 based models to medical and scientific literature \cite{BioMedLM, luo2022biogpt, taylor2022galactica}. 
Recent works such as MedAlpaca \cite{han2023medalpaca} focus on finetuning large scale open-source LLMs such as LLaMA \cite{touvron2023llama, touvron2023llama2} on a set of medical instructions. 
ChatDoctor  \cite{yunxiang2023chatdoctor} enhanced a LLaMA model on 100K clinical Q\&As, incorporating knowledge retrieval capabilities. 
ClinicalCamel \cite{toma2023clinical} introduced question answering data with the Dialogue-Based Knowledge Encoding (DBKE) process, converting PubMed articles and MedQA into questions  and descriptive answers. PMC-LLaMA \cite{wu2023pmc} and Meditron \cite{chen2023meditron} perform pre-training on PubMed content and medical texts, with further refinements on individual MCQA datasets. Med42 \cite{med42} instruction-tuned LLaMA model for medical tasks, though the details of its training remain undisclosed.

Recent years have witnessed progress in Arabic language processing models, including notable models such as AraT5 \cite{nagoudi2021arat5} and AraBART \cite{eddine2022arabart}. Beyond these monolingual frameworks, Arabic has also been incorporated into multilingual systems \cite{workshop2022bloom, muennighoff2022crosslingual}. The recent launch of the Jais model \cite{sengupta2023jais} marked a substantial advancement as a general-purpose bilingual LLM for English and Arabic. Additionally, many of the latest foundational models \cite{touvron2023llama,touvron2023llama2,jiang2024mixtral} demonstrate the ability to reply in Arabic, despite their limited exposure to the language during pre-training. However, these generic models  \cite{sengupta2023jais,touvron2023llama,touvron2023llama2,jiang2024mixtral} lack sufficient medical domain knowledge, making them ill-equipped for medical applications in both Arabic and English languages.  

\begin{table*}[ht]
\centering
\adjustbox{max width=\textwidth}{
\begin{tabular}{ll}
\toprule
\multicolumn{1}{l|}{Dataset} & \multicolumn{1}{c}{Description} \\ \midrule
\multicolumn{2}{l}{\textbf{Multiple-Choice Question Answering (MCQA)}} \\ \midrule
\multicolumn{1}{l|}{PubMedQA \cite{jin2019pubmedqa}} & Closed-answer questions from medical abstracts in PubMed research papers. \\
\multicolumn{1}{l|}{MedMCQA \cite{pal2022medmcqa}} & Multiple-choice questions from the Indian AIIMS and NEET PG medical entrance exams. \\
\multicolumn{1}{l|}{MedQA  \cite{jin2021disease}} & Multiple-choice questions from the USMLE for U.S. medical licensing. \\ \midrule
\multicolumn{2}{l}{\textbf{Question Answering (QA)}} \\ \midrule
\multicolumn{1}{l|}{HealthCareMagic \cite{yunxiang2023chatdoctor}} & Specialist-patient Q\&As crawled from HealthCareMagic. \\
\multicolumn{1}{l|}{iCliniq  \cite{yunxiang2023chatdoctor}} & Physician-patient Q\&As from the iCliniq consultation site. \\
\multicolumn{1}{l|}{Medical Meadow \cite{han2023medalpaca}} & Q\&A pairs from Anki flashcards and Wikidoc’s medical textbooks and Patient Information. \\
\multicolumn{1}{l|}{UMLS \cite{wu2023pmc}} & Q\&As on medical term meanings and entity relationships from knowledge graphs. \\
\multicolumn{1}{l|}{LiveQA \cite{abacha2017overview}} & Annotated consumer health questions from the National Library of Medicine (NLM). \\
\multicolumn{1}{l|}{MedicationQA \cite{abacha2019bridging}} & Annotated consumer medications inquiries from MedlinePlus. \\ \bottomrule
\end{tabular}}
\caption{Summary of collected data sources for multiple-choice (MCQA) and open question answering (QA).}
\label{tab:sources}
\end{table*}

\section{Method}
\noindent\textbf{Overview}: To develop a bilingual healthcare chat assistant, we first introduce a comprehensive bilingual dataset named BiMed1.3M that encompasses  diverse medical interactions such as  multiple-choice question answers (MCQAs), open-ended questions, and multi-turn chat conversations, in both English and Arabic (Sec.~\ref{sec:data}).  This bilingual dataset is created  by first compiling  English instruction, as outlined in Sec. \ref{sec_english}. Subsequently, in Sec.~\ref{sec:arabic}, we describe our semi-automated English-to-Arabic translation pipeline, which enables us to obtain high-quality Arabic medical benchmarks and instruction sets, as detailed in Sec.  \ref{databenchmark}. The generated Arabic instructions are then combined with our English instructions to obtain the BiMed1.3M bilingual dataset. Finally, we present the model and instruction-tuning \mbox{strategy in Sec.  \ref{sec:finetuning}.}

\begin{figure*}[ht]
    \centering
    \begin{minipage}[ht]{.48\textwidth}
        \centering
        \adjustbox{max width=0.85\textwidth}{
        \begin{tabular}{l|ccc} \toprule
           Data & Samples & Avg. Turns & \#Tokens \\ \hline
            QA & 423.8 K & 1.00 & 131.8 K \\
            MCQA & 638.1 K & 1.00 & 342.5 M \\
            Chat & 249.7 K & 4.72 & 158.0 M \\ \hline
            Total & 1311.6 K & 1.71 & 632.3 M \\ \hline
            \end{tabular} }
            \caption{Statistics of the BiMed1.3M across QA, MCQA, and Chat (with more than one turn of exchanges) totaling 623M tokens and 1.3M samples. This bilingual dataset, derived from translating approximately 50\% of the English dataset into Arabic, is 1.5 times larger due to its 2:1 English-to-Arabic content ratio.}
        \label{tab:statistics2}
    \end{minipage}
    \hfill
    \centering
    \begin{minipage}[ht]{.48\textwidth}
        \centering
        \includegraphics[width=\linewidth]{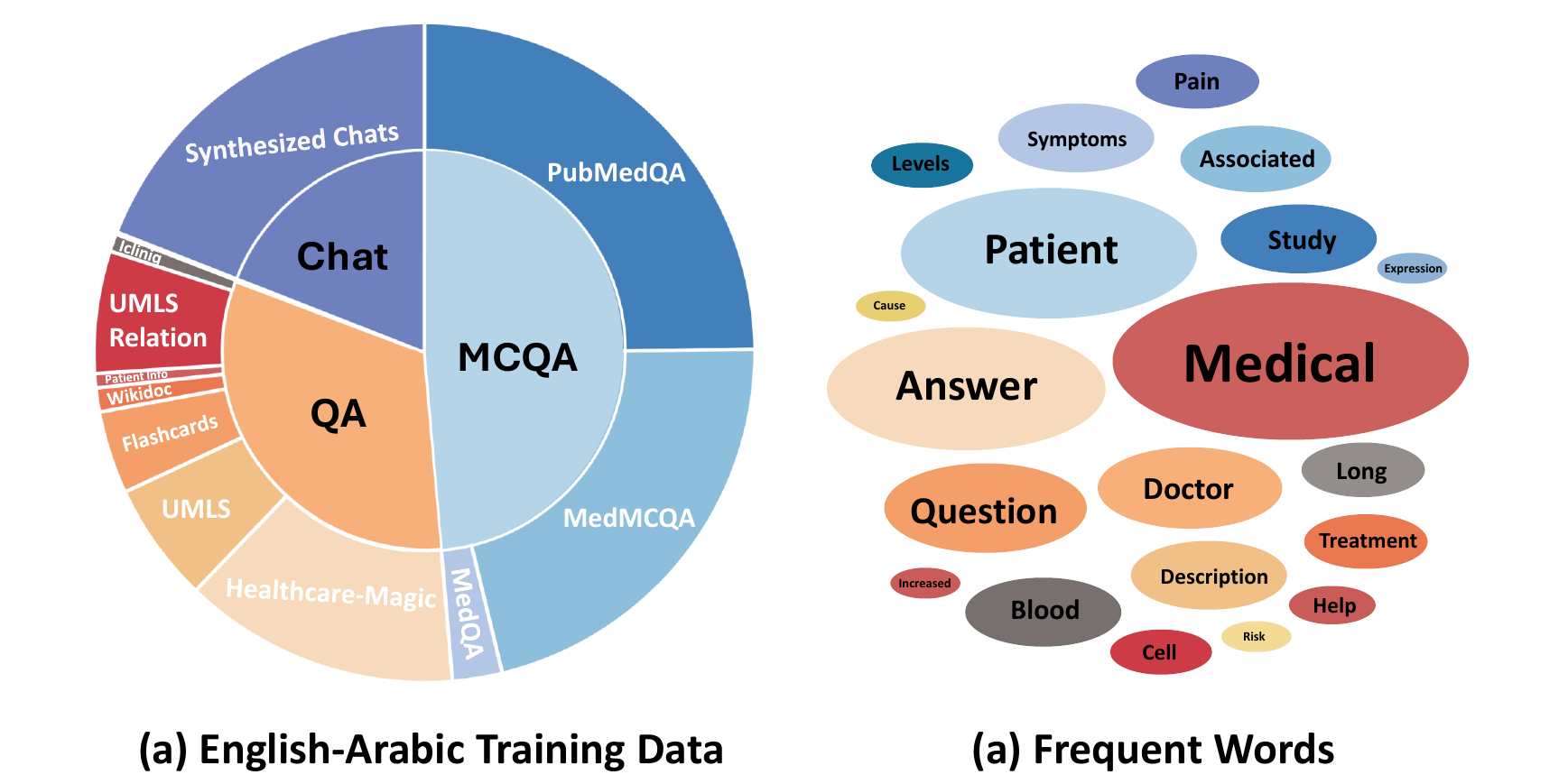}
        \caption{BiMed1.3M Dataset Overview: (a) Data type distribution and (b) Analysis of frequent words.}
        \label{fig:arabic-data}
    \end{minipage}
\end{figure*}

\subsection{BiMed1.3M: Bilingual Dataset with Diverse Medical Interactions}
\label{sec:data}

The growing need for AI-driven medical assistants, proficient across a spectrum of NLP tasks, highlights the necessity for comprehensive datasets. 
In particular, the ability to deliver concise answers is critical for evaluating model performance, yet the functionality for engaging in chat is essential for practical deployment. This holds particular significance in the healthcare domain, where multiple turns of interaction with further inquiries about symptoms, examinations, and pre-existing conditions are often necessary. To this end, we compile our English instruction set as detailed below. 
\subsubsection{Compiling English Instruction Set}
\label{sec_english}
At first, we compile a dataset in English encompassing three types of medical interactions: multiple-choice question answering (MCQA), which focuses on specialized medical knowledge and reasoning for definitive answers; open question answering (QA), that includes real-world consumer questions; and multi-turn chat conversations for dynamic exchanges.
For the first two categories, we combine various existing sources into a unified collection of question-answer pairs, as detailed in Table \ref{tab:sources}. For the multi-turn chat component, we generate realistic conversations grounded on MCQA scenarios, as described below.\\

\noindent\textbf{MCQA-grounded Multi-turn Chat Generation:}
To generate realistic and engaging multi-turn conversations that are medically accurate, we leverage ChatGPT's conversational flow together with publicly available medical MCQAs (multiple-choice question answers).  Utilizing ChatGPT, we simulate potential doctor-patient dialogues based on MCQAs. The chat creation process involves using ChatGPT to draft messages for both the user and the AI assistant in a dialogue format. Inputs from the PubmedQA, MedQA, and MedMCQA datasets are provided with specific directives to generate multi-turn discussions relevant to the medical  question answer pairs (MCQA).  A prompt template is employed to define the format and criteria alongside the input question, generating conversation transcripts for both parties (doctor and patient) until a logical conclusion is reached.

The MCQA-grounded multi-turn interactions produced by our approach are generally thorough and insightful, covering aspects like symptom inquiries and further information requests. Based on this approach, we produce more than 200,000 high-quality multi-turn medical dialogues, each linked to a specific MCQA, collectively comprising over 74 million tokens.

Following the generation of multi-turn dialogues, our dataset comprehensively encompasses multiple-choice question answering (MCQA), open question answering (QA), and dynamic chat conversations (Chats)—totaling over 860,000 instruction-tuning data in English. Moreover, we leverage various publicly available medical benchmark datasets in English to evaluate the performance of our model, as elaborated in Sec.~\ref{sec:eval-datasets}. 
Next, we detail our semi-automated iterative translation pipeline to generate medical evaluation benchmarks and instruction sets tailored for the resource-constrained Arabic language.

\begin{figure*}[htbp]
\centering
\includegraphics[width=\textwidth]{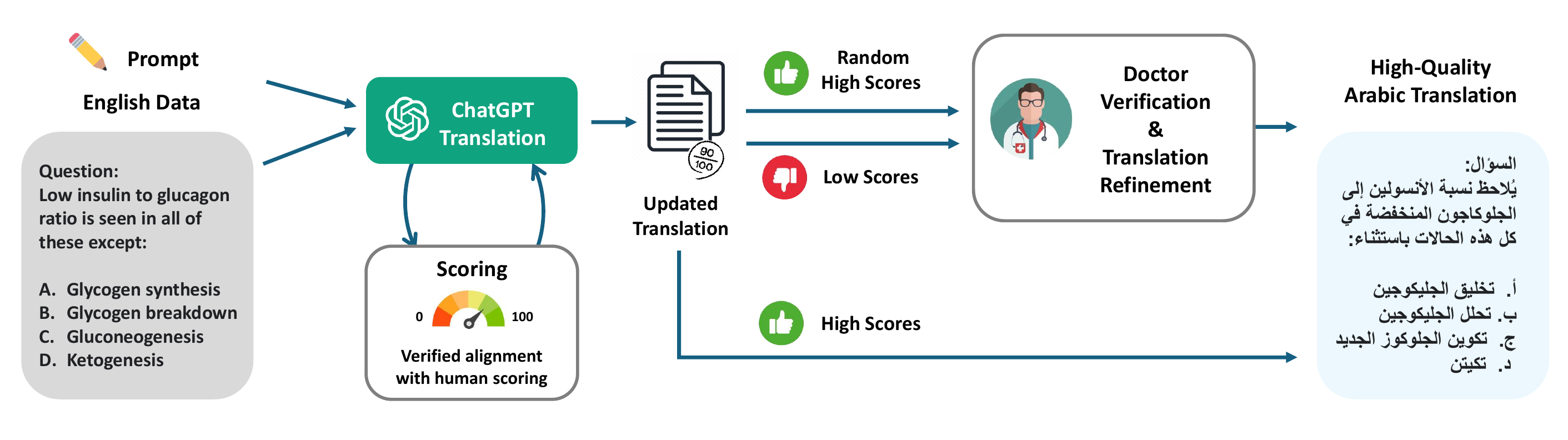} 
\caption{Overview of the proposed semi-automated, iterative translation pipeline featuring human alignment. The process involves iterative translation and score optimization utilizing LLMs (such as ChatGPT) and manual refinement by professionals for samples with low scores along with a random selection of high-scoring samples.}
\label{fig:benchmarks}
\end{figure*}

\subsubsection{Semi-Automated Iterative Translation}
\label{sec:arabic}

Our semi-automated iterative translation pipeline is shown in Fig.~\ref{fig:benchmarks}. 
At first, we perform English to Arabic translation using ChatGPT, which delivers high-quality translations by considering the entire input English text as context. Subsequently, we task ChatGPT with evaluating the quality of the translated Arabic text relative to its original English version. We assign scores ranging from 0 to 100, directly proportional to the translation quality. This scoring  aims to quantitatively assess the fidelity and clarity of translations, ensuring the accurate preservation of technical terminology and specifics. Additionally, we verify the alignment of these scores  to human preferences with the assistance of native Arabic speakers.

 For translations that score below a predefined threshold, we initiate a refinement process. This involves providing ChatGPT with the original English text, along with the current translations and their scores as feedback. We prompt ChatGPT to update the translation to ensure optimal consistency and alignment with the English originals. This feedback loop iteratively improves the translation quality through successive revisions. 
 
 However, we observed that certain samples with initially low scores faced limitations in enhancing their scores and translation quality through this automated procedure. To address this, we subject all translations with scores below a threshold to a thorough \textit{manual verification process performed by a medical professional fluent in Arabic}. To ensure 
high quality, we additionally sample a random subset of translations with high scores for professional review.  Our iterative translation process  ensures that translations adhere to rigorous academic and clinical standards. 

\subsubsection{Bilingual Benchmark \& Instruction Set}
\label{databenchmark}

\noindent\textbf{Creation of Medical Benchmark:} To evaluate the accuracy and applicability of Arabic medical AI models, we translate the English medical evaluation benchmarks (detailed in Sec. \ref{sec:eval-datasets}) into Arabic using the aforementioned iterative translation procedure. By providing a high-quality Arabic medical benchmark aligned with its English counterpart, we aim to establish a fundamental step in bridging the linguistic divide in model evaluation and comparison, thereby offering a valuable asset for further research \footnote{The Arabic benchmark is available for download at the Hugging Face repository \url{https://huggingface.co/BiMediX}   .}.

These Arabic benchmarks are combined with the original English evaluation benchmarks to create an English-Arabic bilingual benchmark. This allows us to assess the linguistic capabilities of our bilingual model as well as its medical domain knowledge.

\noindent\textbf{Bilingual Instruction Set:} We translate 444,995 English samples into Arabic, covering all three types of medical interactions. Following \cite{sengupta2023jais}, we adopted a bilingual approach, mixing Arabic and English in a 1:2 ratio. This approach led to the creation of an extensive bilingual instruction tuning dataset that integrates both languages. Consequently, we compile an English-Arabic bilingual dataset named BiMed1.3M, 1.5 times larger than our English counterpart, comprising more than 1,311,000 samples. Further details are illustrated in Tab. \ref{tab:statistics2} and Fig. \ref{fig:arabic-data}. 

Next  we introduce our model and its bilingual medical instruction tuning.

\subsection{Bilingual Medical Instruction Tuning of Mixture of Experts LLM}
\label{sec:finetuning}

Recently,  Mixtral \cite{jiang2024mixtral}  introduced a  Mixture of Experts (MoE) based architecture that achieves impressive performance on generic NLP benchmarks in English.  It enables substantial scaling of model size within the same computational budget as traditional dense models. Unlike traditional dense feed-forward network layers, Mixtral employs a set number of "experts" in a sparse manner, replacing the FFN layers in dense networks. Additionally, a gate network or router is employed to direct input tokens to appropriate experts based on learned parameters. 
While Mixtral \cite{jiang2024mixtral} offers advantages such as improved inference times over dense models, it faces challenges in  Arabic-English bilingual medical applications  due to limited medical domain knowledge and Arabic language capabilities. To address these challenges, we perform Arabic-English bilingual medical instruction tuning  of  Mixtral MoE architecture using our BiMed1.3M dataset.  
Note that performing conventional supervised fine-tuning of  Mixtral \cite{jiang2024mixtral} requires substantial computational resources and often encounters challenges in carefully tuning the expert layers and the routing network, leading to generalization issues. To this end, we employ Parameter-Efficient Finetuning (PEFT) techniques \cite{hu2021lora, li2021prefix, dettmers2023qlora} to adapt the pre-trained Mixtral model \cite{jiang2024mixtral} with minimal computational resources while ensuring high performance and adaptability. In our approach, we use QLoRA-based PEFT training.  Specifically, we attach the QLoRA adapters to the decoder layers of \cite{jiang2024mixtral}, including the experts and the routing network, to repurpose the Mixture of Experts architecture as a bilingual medical chat assistant by utilizing the proposed BiMed1.3M dataset for instruction tuning. 

Moreover, to align with the conversational style  in our BiMed1.3M dataset, for both English and Arabic instructions, we adopt the Vicuna \cite{chiang2023vicuna} data format {"conversations": [{"from": "human/AI," "value": "..."}]}, calculating loss exclusively for the AI's responses within the dialogue script. Unlike alternative approaches such as \cite{wu2023pmc}, we refrain from conducting additional medical pre-training before instruction-tuning it on our medical dataset.

In summary, through careful instruction tuning of the routing network and experts in the Mixtral decoder using our BiMed1.3M dataset, we achieve enhanced medical domain capabilities in both Arabic and English languages. This includes proficiency in multiple-choice question answering, open-ended question answering, and realistic multi-turn question answering between our BiMediX model and the user (e.g., patient), making it suitable for deployment as a bilingual medical chatbot.

\section{Experiments}

\subsection{Experimental Settings}
Our BiMediX is built upon Mixtral-8x7B \cite{jiang2024mixtral} base network, a sparse mixture of experts language model. This model extends the Mistral-7B architecture \cite{jiang2023mistral}, featuring a unique design with each layer hosting eight "expert" feedforward blocks. A router network determines which two experts will process each token, merging their outputs. Consequently, it handles 47 billion parameters, with fewer than 13 billion active during inference.
This model features a hidden state dimension of 14,336, a context window capable of accommodating 32,768 tokens, 32 layers, 32 attention heads, and a vocabulary size of 32,000.

For fine-tuning Mixtral, we utilize QLoRA, a low-rank adaptation technique, by incorporating a set of learnable low-rank adapter weights into both the experts and the routing network. This adaptation applies to \(q\), \(k\), and \(v\), setting the rank to 128 and the \(\alpha\) parameter to 64. Consequently, this approach results in the training of approximately \textit{4\% of the original parameters}.
Our bilingual model underwent training on roughly 632 million tokens sourced from the Arabic-English corpus, including 288 million tokens that are exclusively for English.
The training hyper-parameters included an effective batch size of 16 with two gradient accumulation steps and utilized the AdamW optimizer with a learning rate of 0.0002, cosine learning rate schedule, and ten warmup steps. 
The model implementation leveraged PyTorch \cite{paszke2019pytorch}, along with the Deepspeed and ZeRO \cite{rasley2020deepspeed, rajbhandari2020zero} packages, with gradient checkpointing \cite{chen2016training}. Our bilingual BiMedix is trained for two epochs with eight A100 (80GB) GPUs, requiring only 35 hours for the training. 

\subsection{Quantitive Evaluation}

In the literature, evaluating medical language models predominantly involves multiple-choice question-answering tasks, with accuracy as the performance metric. We employed the EleutherAI evaluation framework\footnote{\url{https://github.com/EleutherAI/lm-evaluation-harness}, is the backend for Hugging Face Open LLM Leaderboard.}.

\subsubsection{Medical Benchmarks}
\label{sec:eval-datasets}

Our analysis incorporates multiple prominent benchmarks in medical multiple-choice question-answering. Among these datasets is MedQA, which is sourced directly from the National Board of Medical Examiners (NBME), the entity tasked with overseeing the assessments necessary for U.S. medical licensure.
    
    \noindent(\textbf{i) PubMedQA} \cite{jin2019pubmedqa} is a question-answering dataset derived from biomedical research papers on PubMed. Given a question sourced from the title of a research paper and a context from the abstract, the task is to provide an answer in the form of 'yes', 'no', or 'maybe', mirroring the conclusion of the paper.
    Building on prior research, our examination focuses on the PQA-L subset of PubMedQA, which is comprised of manually annotated question-answer (QA) pairs sourced from PubMed. Answering these questions with accuracy requires thorough reasoning over the biomedical contexts and quantitative data analysis.
    
    \noindent\textbf{(ii) MedMCQA} \cite{pal2022medmcqa} is a multiple-choice dataset constructed from questions featured in the Indian AIIMS and NEET PG medical entrance exams. It covers a broad spectrum of medical subjects, evaluating both professional domain knowledge and language comprehension. The dataset includes a test set of 4,183 questions, each with four options.
    
    \noindent\textbf{(iii) MedQA} \cite{jin2021disease} is a dataset comprising multiple-choice questions from medical board examinations in the US, Mainland China, and Taiwan. It features two types of questions: concise sentences to assess specific knowledge and extended paragraphs that detail a patient's condition. Our analysis concentrates on the English portion (USMLE), containing 1,273 samples for testing purposes. Engaging with this benchmark involves multiple stages of reasoning and evidence retrieval.

    \noindent\textbf{(iv) The Medical MMLU} (Massive Multitask Language Understanding) is  a group of six datasets that compiles 1,089 test questions with \mbox{four options} related to different medical subjects.  The six MMLU datasets are  Clinical Knowledge (Cli-KG),  College Biology (C-Bio),  College Medicine (C-Med),  Medical Genetics (Med-Gen),  Professional Medicine (Pro-Med) \& Anatomy (Ana).

    All 7045 questions from the above English datasets are translated into Arabic using our semi-automated pipeline to create our Arabic and Bilingual evaluation benchmarks. 
   
    \begin{table*}[!h]
    \centering
    \resizebox{\textwidth}{!}{
    \begin{tabular}{@{}lcccccccccccc@{}}
    \toprule
     & \multicolumn{6}{c}{\textbf{MMLU}} & \multicolumn{1}{c}{\textbf{MedMCQA}} & \multicolumn{1}{c}{\textbf{MedQA}}  & \multicolumn{1}{c}{\textbf{PubmedQA}} & \multicolumn{1}{c}{\textbf{AVG}} \\ \cmidrule(lr){2-7}
     \multirow{-2}{*}{\textbf{Model}} & \textbf{Cli-KG} & \textbf{C-Bio} & \textbf{C-Med} & \textbf{Med-Gen} & \textbf{Pro-Med} & \textbf{Ana} & & & &  \\ \midrule
    Jais-30B   \cite{sengupta2023jais}      & 57.4 & 55.2 & 46.2 & 55.0 & 46.0 & 48.9 & 40.2 & 31.0 & 75.5 &  50.6 \\
    Mixtral-8x7B  \cite{jiang2024mixtral}  & 59.1 & 57.6 & 52.6 & 59.5 & 53.3 & 54.4 & 43.2 & 40.6 & 74.7 &  55.0 \\
    \rowcolor{blue!10} \textbf{BiMediX (Bilingual)} & \textbf{70.6} & \textbf{72.2} & \textbf{59.3} & \textbf{74.0} & \textbf{64.2} & \textbf{59.6} & \textbf{55.8} & \textbf{54.0} & \textbf{78.6} & \textbf{65.4} \\ \bottomrule
    \end{tabular}
    }
    \caption{Performance of BiMediX on the \textbf{Bilingual benchmark}.}
    \label{tab:bi-results}
    \end{table*}

    \begin{table*}[!h]
    \centering
    \resizebox{\textwidth}{!}{
    \begin{tabular}{@{}lcccccccccccc@{}}
    \toprule
     & \multicolumn{6}{c}{\textbf{MMLU}} & \multicolumn{1}{c}{\textbf{MedMCQA}} & \multicolumn{1}{c}{\textbf{MedQA}}  & \multicolumn{1}{c}{\textbf{PubmedQA}} & \multicolumn{1}{c}{\textbf{AVG}} \\ \cmidrule(lr){2-7}
     \multirow{-2}{*}{\textbf{Model}} & \textbf{Cli-KG} & \textbf{C-Bio} & \textbf{C-Med} & \textbf{Med-Gen} & \textbf{Pro-Med} & \textbf{Ana} & & & &  \\ \midrule
    Jais-30B \cite{sengupta2023jais}      & 52.1 & 50.7 & 40.5 & 49.0 & 39.3 & 43.0 & 37.0 & 28.8 & 74.6 & 46.1 \\
    BiMediX (Arabic) & 60.0 & 54.9 & \textbf{55.5} & 58.0 & \textbf{58.1} & 49.6 & 46.0 & 40.2 & 76.6 & 55.4 \\
    \rowcolor{blue!10} \textbf{BiMediX (Bilingual)} & \textbf{63.8} & \textbf{57.6} & 52.6 & \textbf{64.0} & 52.9 & \textbf{50.4} & \textbf{49.1} & \textbf{47.3}  & \textbf{78.4} & \textbf{56.5} \\ \bottomrule
    \end{tabular}
    }
    \caption{Performance of BiMediX on the \textbf{Arabic benchmark}.}
    \label{tab:ara-results}
    \end{table*}

    \begin{table*}[!h]
    \centering
    \resizebox{\textwidth}{!}{
    \begin{tabular}{@{}lcccccccccccc@{}}
    \toprule
    & \multicolumn{6}{c}{\textbf{MMLU}} & \multicolumn{1}{c}{\textbf{MedMCQA}} & \multicolumn{1}{c}{\textbf{MedQA}} & \multicolumn{1}{c}{\textbf{PubmedQA}} &  \multicolumn{1}{c}{\textbf{AVG}} \\ \cmidrule(lr){2-7}
    \multirow{-2}{*}{\textbf{Model}} & \textbf{Cli-KG} & \textbf{C-Bio} & \textbf{C-Med} & \textbf{Med-Gen} & \textbf{Pro-Med} & \textbf{Ana} & & & &  \\
    \midrule
    PMC-LLaMA-13B  \cite{wu2023pmc}     & 63.0 & 59.7 & 52.6 & 70.0 & 64.3 & 61.5 & 50.5 & 47.2 & 75.6 &  60.5 \\
    Med42-70B     \cite{med42}      & 75.9 & 84.0 & 69.9 & 83.0 & 78.7 & 64.4 & 61.9 & 61.3 & 77.2 &  72.9 \\
    Clinical Camel-70B \cite{toma2023clinical}  & 69.8 & 79.2 & 67.0 & 69.0 & 71.3 & 62.2 & 47.0 & 53.4 & 74.3 &  65.9 \\
    Meditron-70B  \cite{chen2023meditron}      & 72.3 & 82.5 & 62.8 & 77.8 & 77.9 & 62.7 & \textbf{65.1} & 60.7 & 80.0 &  71.3 \\
     \rowcolor{blue!10} \textbf{BiMediX} & \textbf{78.9} & \textbf{86.1} & \textbf{68.2} & \textbf{85.0} & \textbf{80.5} & \textbf{74.1} & 62.7 & \textbf{62.8}  & \textbf{80.2} & \textbf{75.4} \\ \bottomrule
    \end{tabular}
    }
    \caption{Performance of BiMediX on the \textbf{English benchmark}.}
    \label{tab:eng-results} 
    \end{table*}

    \begin{table}[ht]
    \centering
    \setlength{\tabcolsep}{3pt}
    \resizebox{\columnwidth}{!}{
    \begin{tabular}{@{}lccc@{}}
    \toprule
    \multicolumn{1}{c}{\textbf{Model}} & \textbf{Active Param} & \textbf{Latency} & \textbf{Tokens/sec} \\ \midrule
    PMC-LLaMA-13B \cite{wu2023pmc} & 13B & 4.1 s & 124.1 \\
    Med42-70B \cite{med42} & 70B & 24.5 s & 20.9 \\
    Meditron-70B \cite{chen2023meditron} & 70B & 24.5 s & 20.9 \\
    Jais-30B \cite{sengupta2023jais} & 30B & 14.0 s & 36.5 \\
    \rowcolor{blue!10} \textbf{BiMediX} & 13B & \textbf{2.8 s} & \textbf{180.6} \\ \bottomrule
    \end{tabular}
    }
\caption{Model statistics comparing memory size and inference speed, computed with \cite{llm_analysis_2023} on GPU A100-80GB.}. 
\label{tab:model-deails}
\end{table}

\textbf{}
\subsubsection{Results}
\noindent{\textbf{Bilingual Evaluation:}} Here, we assess our BiMediX on  Arabic-English bilingual evaluation benchmark  derived from evaluating results in both languages. Tab.~\ref{tab:bi-results} shows our BiMediX
results against the base model, Mixtral-8x7B (not fine-tuned) and Jais-30B. The latter is a larger model designed specifically for the Arabic language and capable of functioning in both English and Arabic.  BiMediX demonstrates superior performance across all benchmarks, achieving accuracy that is, on average, more than 10 and 15 points higher, respectively, when compared to the baseline model and Jais-30B. This achievement underscores the substantial value of our proposed BiMed1.3M dataset and its unmatched effectiveness and adaptability in addressing medical queries within a bilingual framework. \\
\noindent\textbf{Arabic Benchmark:} We test BiMediX using our Arabic benchmark, comparing its performance with that of Jais-30B. In Table \ref{tab:ara-results}, we present the findings for Jais and our BiMediX in two configurations: one pre-trained exclusively on Arabic content (Arabic) and the other with bilingual data (Bilingual). Our bilingual model outperforms in all categories within the Arabic context, underscoring that the integration of both types of training data significantly enhances the understanding and processing of medical information in an Arabic setting.\\
\noindent\textbf{English Benchmark:} We evaluate the quantitative performance of our English model against previous state-of-the-art English medical models. As illustrated in Table \ref{tab:eng-results}, BiMediX exhibits outstanding performance across all subsets, securing the highest average scores among all models assessed. 
When compared to  Clinical Camel-70B model, our  BiMediX exhibits around 10\% average performance gain.  Furthermore, it also outperforms Meditron-70B in almost every subset except for MedMCQA. Notably, our model's success comes without the need for separate fine-tuning on the training set for each evaluation benchmark, in contrast to Meditron, which employs individual fine-tuning for each evaluation benchmark to achieve favorable outcomes. 
This demonstrates our model's versatility in handling various medical interactions simultaneously.
Moreover, our method outperforms Med42 by an average  gain of 2.5\%
Furthermore, our model is more efficient in terms of prediction speed (latency and tokens per second) than all other models considered, as detailed in Table \ref{tab:model-deails}.     Additional qualitative examples are provided in the Appendix. 

 \section{Conclusion}

Our work introduces the first bilingual medical mixture of experts (LLM) BiMediX designed to facilitate comprehensive medical interactions, including conversations, MCQA, and QA in both English and Arabic languages. The implementation of a semi-automated translation pipeline, coupled with human refinement, ensures the high quality of English-to-Arabic translations. BiMediX demonstrates superior performance over existing models such as Med42 and Meditron in English-based medical evaluation benchmarks, and significantly outperforms the generic bilingual LLM, Jais-30B, in Arabic medical and bilingual evaluations. The introduction of a novel evaluation benchmark and the extensive BiMed1.3M instruction set, encompassing 1.3 million diverse medical interactions, including chats, further highlights the significance of our contributions.

\section{Limitations}
Despite its overall improvement, BiMediX, like other language models, may experience issues such as hallucinations, toxicity, and stereotypes. These issues stem from both the inherited limitations of the base models and the nature of the pre-training data. While we have conducted automatic and qualitative evaluations with medical professionals, we acknowledge that our model's medical diagnoses and recommendations may not always be accurate. Extensive human evaluation is more reliable but costlier and time-consuming. The exploration of alternative solutions remains an important focus for ongoing research. Currently, our models lack explicit mechanisms to curb undesirable behaviors. Future research will focus on enhancing security and alignment strategies \cite{christiano2017deep, bai2022constitutional, rafailov2023direct}. On a brighter note, we believe that releasing our weight could contribute to investigating and mitigating risks tied to application uses.

\section{Safety and Ethical Implications}
We recognize the significant societal impact of this technology, emphasizing ethical considerations and transparency. 
This release, intended for research, is not ready for clinical or commercial use. Further research is needed to ensure safety and accuracy in clinical settings and to prevent patient harm. Collaboration with patients, medical professionals, and ethicists is crucial for ethical oversight. 

\bibliography{acl_latex}

\newpage

\label{app:examples}
\begin{figure*}[ht]
  \centering
    \includegraphics[width=0.8\linewidth]{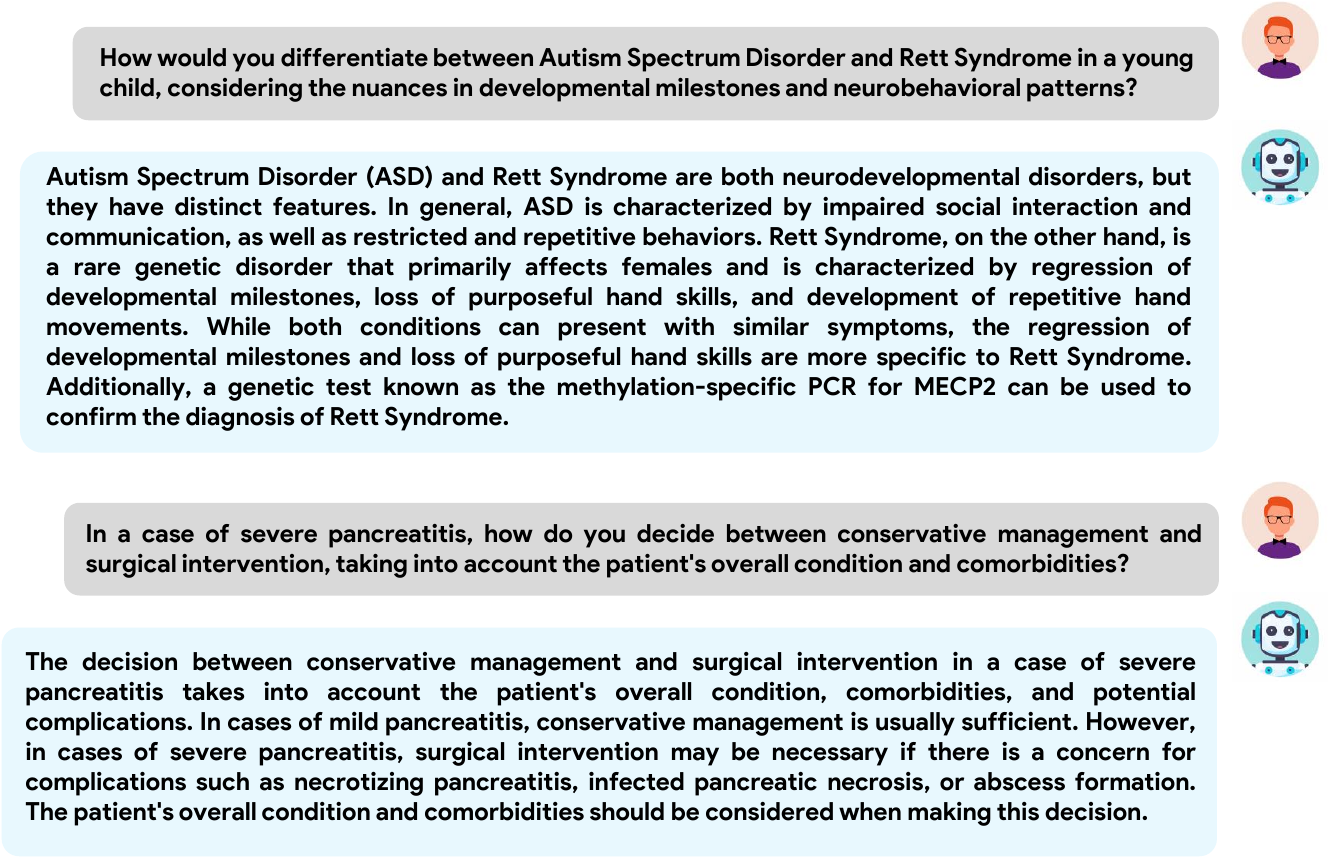}
    \includegraphics[width=0.8\linewidth]{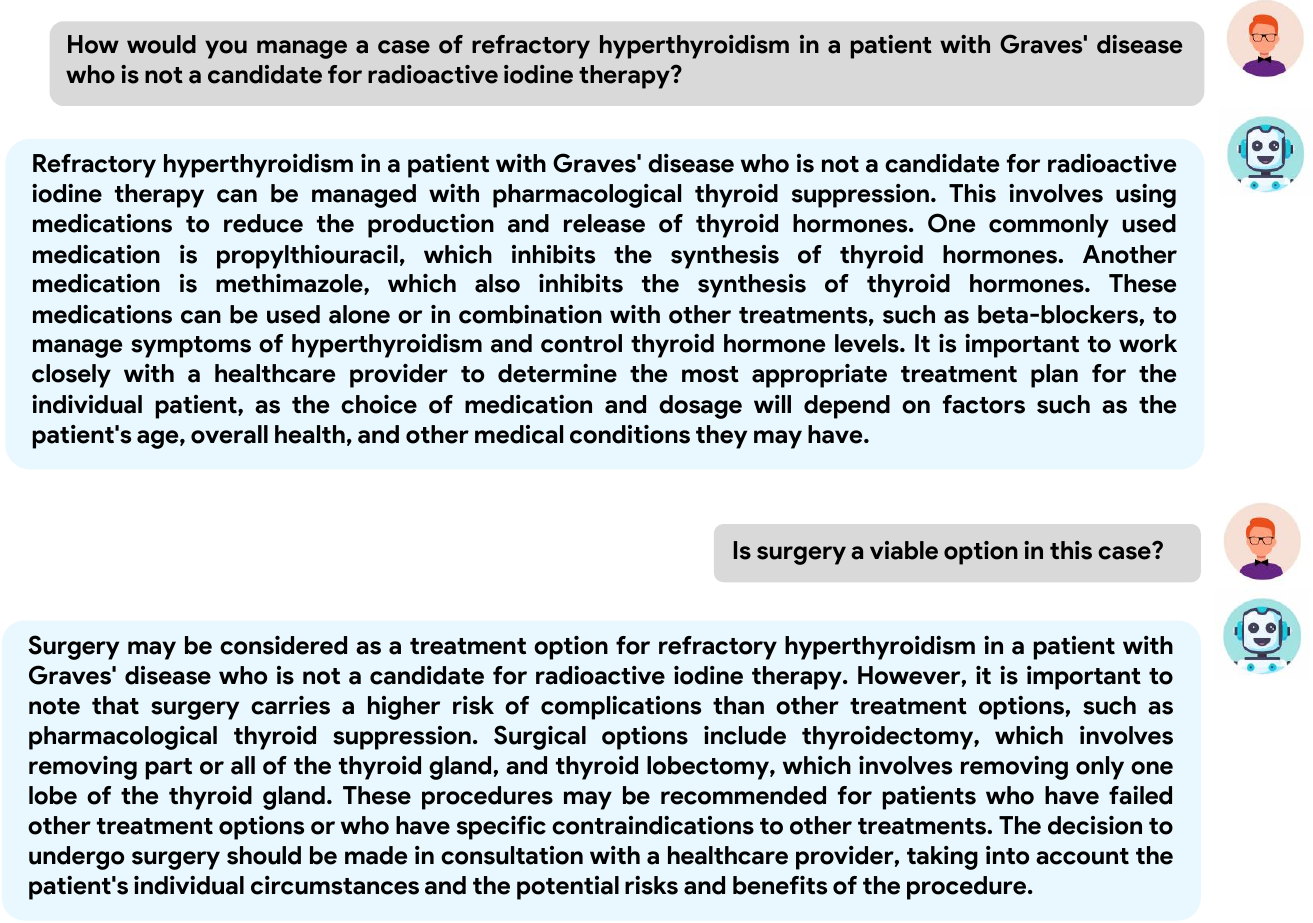}
    \caption{\textbf{Qualitative Examples of our BiMediX.}}
    \label{fig:example2}
\end{figure*}

\begin{figure*}[ht]
  \centering
    \includegraphics[width=0.8\linewidth]{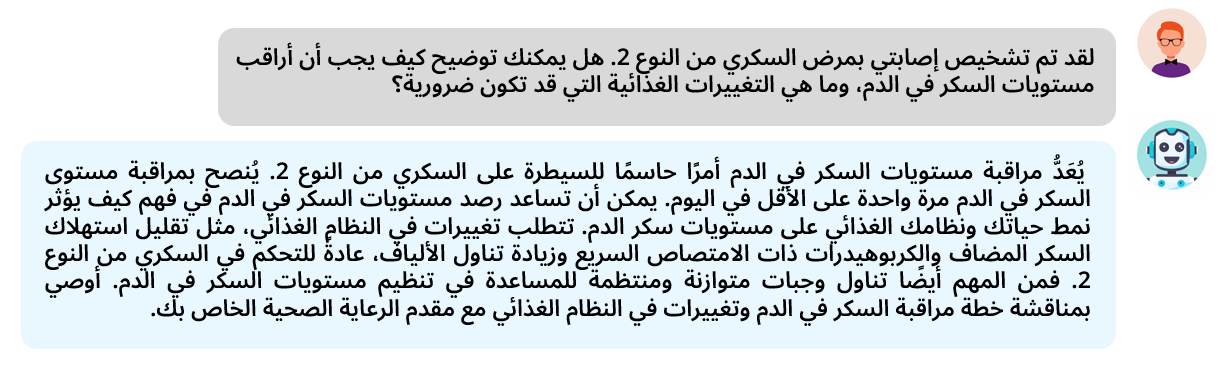}
    \includegraphics[width=0.8\linewidth]{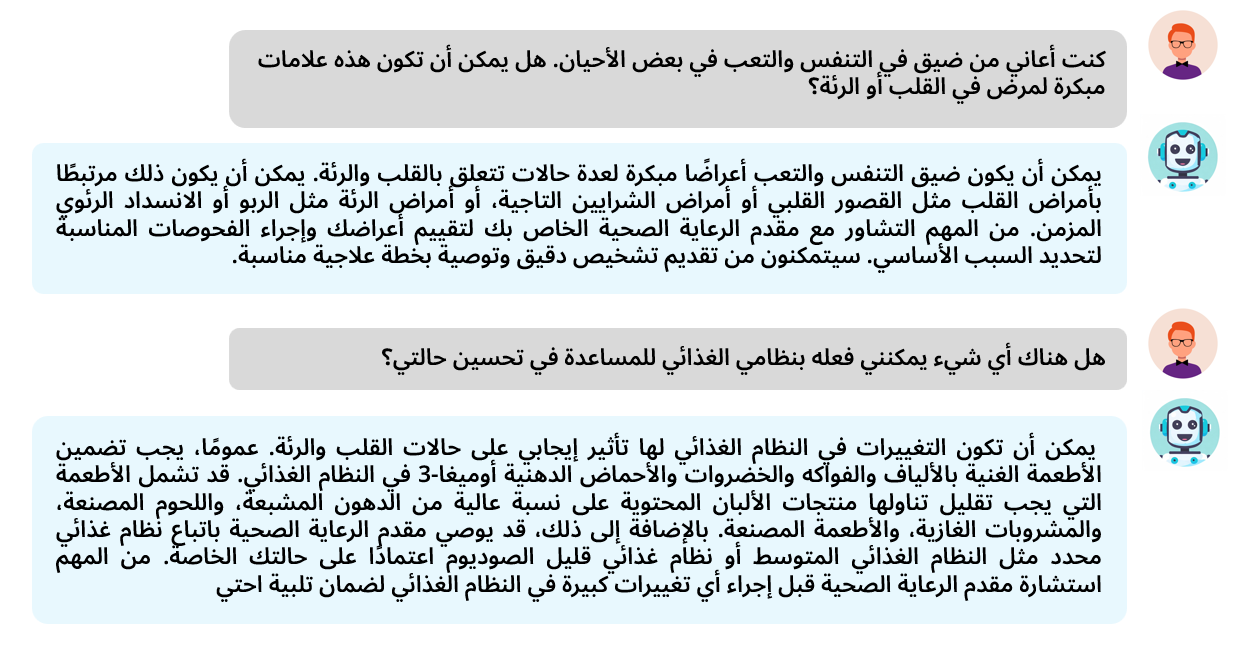}
    \caption{\textbf{Qualitative Examples of our BiMediX, which is translated to English as follows:} \\ \textbf{User:} \textit{I have been diagnosed with type 2 diabetes. Can you explain how I should monitor my blood sugar levels, and what dietary changes might be necessary?} \textbf{BiMediX:} \textit{Monitoring blood sugar levels is crucial to controlling type 2 diabetes. It is recommended to monitor your blood sugar level at least once a day. Monitoring your blood sugar levels can help you understand how your lifestyle and diet affect your blood sugar levels. It usually requires dietary changes, such as reducing consumption of added sugar and fast-absorbing carbohydrates and increasing fiber intake, to control type 2 diabetes. It is also important to eat regular, balanced meals to help regulate blood sugar levels. I recommend discussing a blood sugar monitoring plan and diet changes with your healthcare provider.}
    \\ \textbf{User:} \textit{I've been experiencing occasional shortness of breath and fatigue. Could these be early signs of a heart or lung condition?} \textbf{BiMediX:} 
 \textit{Shortness of breath and fatigue can be early symptoms of several heart and lung conditions. This can be related to heart disease such as heart failure or coronary artery disease, or lung disease such as asthma or chronic obstructive pulmonary disease. It is important to consult with your healthcare provider to evaluate your symptoms and perform appropriate tests to determine the underlying cause. They will be able to provide an accurate diagnosis and recommend an appropriate treatment plan.}
    \\ \textbf{User:} \textit{Is there anything I can do with my diet to help improve my situation ?} \textbf{BiMediX:} \textit{Dietary changes can have a positive impact on heart and lung conditions. Generally, foods rich in fiber, fruits, vegetables and omega-3 fatty acids should be included in the diet. Foods to reduce your intake may include dairy products high in saturated fat, processed meats, soft drinks, and processed foods. In addition, your healthcare provider may recommend a specific diet such as a moderate-moderate diet or a low-sodium diet depending on your specific condition. It is important to consult your healthcare provider before making any major dietary changes to ensure your needs are met.
}}
    \label{fig:example3}
\end{figure*}

\end{document}